\pdfoutput=1
\documentclass[letterpaper, 10 pt, conference]{ieeeconf} 

\IEEEoverridecommandlockouts               

\usepackage{geometry}
\usepackage{graphicx} 
\usepackage{cancel}

\usepackage[dvipsnames]{xcolor}
\usepackage{algorithm}
\usepackage{algorithmic} 
\usepackage{optidef}
\usepackage{mathtools, cuted}

\makeatletter
\newcommand\fs@norules{\def\@fs@cfont{\bfseries}\let\@fs@capt\floatc@ruled
  \def\@fs@pre{}%
  \def\@fs@post{}%
  \def\@fs@mid{\kern3pt}%
  \let\@fs@iftopcapt\iftrue}
\makeatother
\floatstyle{norules}
\restylefloat{algorithm}

\newcommand{\norm}[1]{\left\lVert#1\right\rVert}

\usepackage{amsmath,amsfonts,amssymb}
\usepackage{multicol} 		
\usepackage{comment}
\usepackage[dvips,frame,rotate,import]{xy} 

\def\doubleunderline#1{\underline{\underline{#1}}}
\title{\LARGE \bf
Control and Dynamic Motion Planning for a Hybrid Air-Underwater Quadrotor: Minimizing Energy Use in a Flooded Cave Environment
}

\author{Ilya Semenov$^{1}$, Robert Brown$^{1}$, Michael Otte$^{2}$
\thanks{$^{1}$Ilya Semenov and Robert Brown, Aerospace Engineering, Alfred Gessow Rotorcraft Center, University of Maryland, College Park
    {\tt\small isemenov@umd.edu} and {\tt\small rbrown36@terpmail.umd.edu}}%
\thanks{$^{2}$Michael Otte, Aerospace Engineering, University of Maryland, College Park
    {\tt\small otte@umd.edu}}%
}

\begin{document}

\maketitle
\thispagestyle{empty}
\pagestyle{empty}

\begin{abstract}

We present a dynamic path planning algorithm to navigate an amphibious rotor craft through a concave time-invariant obstacle field while attempting to minimize energy usage. We create a nonlinear quaternion state model that represents the rotor craft dynamics above and below the water. The 6 degree of freedom dynamics used within a layered architecture to generate motion paths for the vehicle to follow and the required control inputs. The rotor craft has a 3 dimensional map of its surroundings that is updated via limited range onboard sensor readings within the current medium (air or water). Path planning is done via PRM and D* Lite. 


\end{abstract}


\section{Introduction} \label{s:intro}

\begin{figure}[t!]
  \begin{minipage}{4.2cm}
	\includegraphics[width=4cm, trim=0 300 0 100, clip=true]{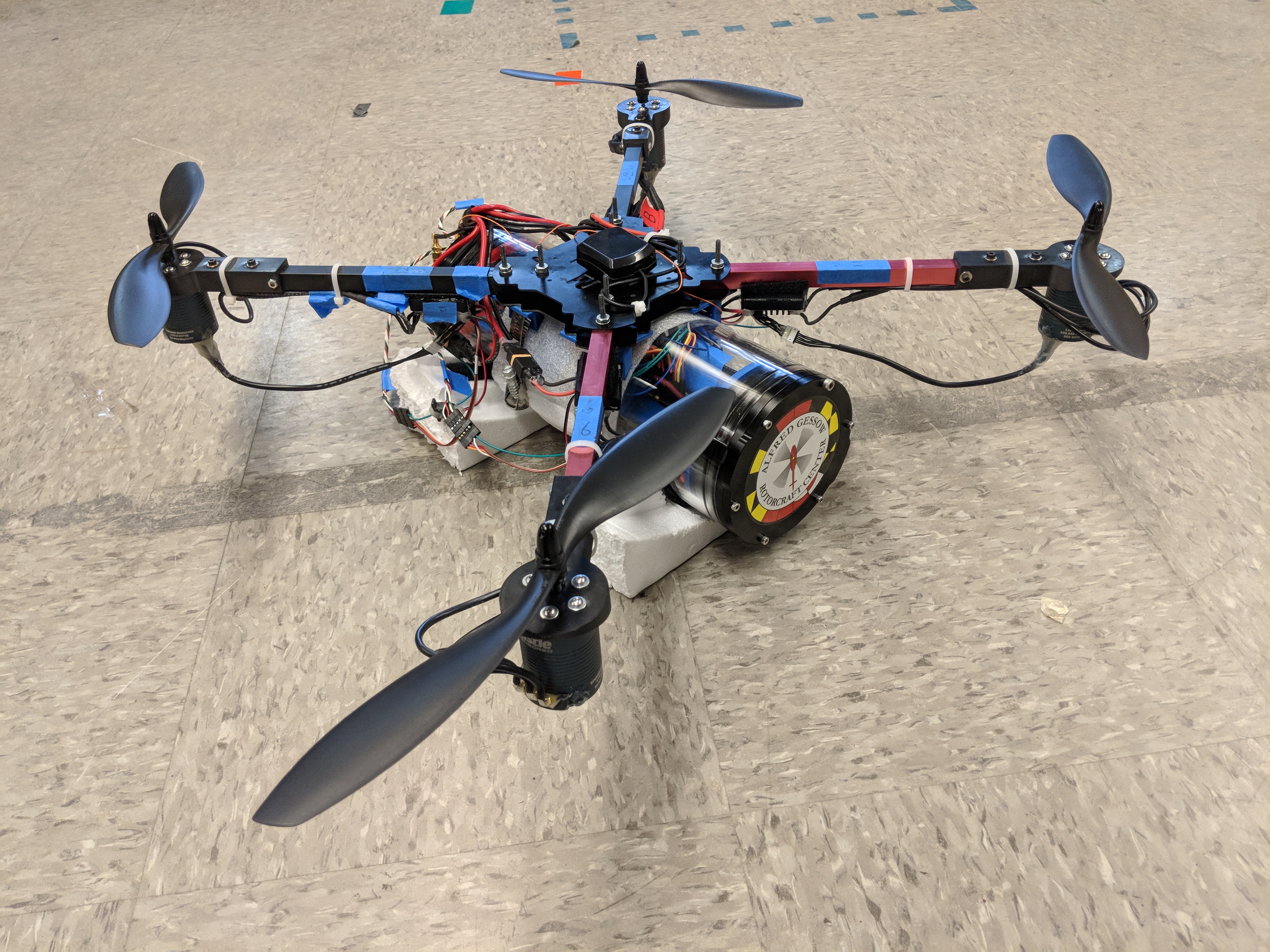}
  \end{minipage}
  \begin{minipage}{4.1cm}
	\caption{The Autonomous Quad With Underwater Ability (AQWUA) vehicle is an amphibious vehicle capable of traveling through both air and water.}
    \label{fig:aqwua}.
  \end{minipage}
\end{figure}

One of the last unexplored frontiers on Earth is below the water surface. As society's use of, impact on, and interaction with Earth's bodies of water increases, so too will the necessity for a complete understanding of the marine environment. In order to further this understanding, an amphibious quad-rotor vehicle shown in Fig. \ref{fig:aqwua} has been developed \cite{ilya}. The \textit{Autonomous Quad With Underwater Ability} (AQWUA) can transition between air and underwater to perform missions within the neighborhood of the water surface. 

This type of vehicle provides many advantages over traditional quad-rotors, submersibles, and other fixed wing and multi-rotor hybrid vehicles. These strengths are emphasized in cave exploration. Many caves have flooded sections as well as in-air sections, meaning only a maneuverable hybrid vehicle may navigate them. Due to a lack of communication, autonomous operation of the vehicle is necessary, and thus the need to solve the path planning problem in an unknown environment arises.

\subsection{Novelty}
This paper focuses on the hybrid air and underwater motion planning problem of the AQWUA vehicle, and considers the scenario of traveling through a partially submerged cave system. Motion planning for an air-underwater hybrid vehicle has not been previously explored to the authors' knowledge.

{\bf The key contributions of this paper are: the formulation of an air-underwater trajectory controller, a motion planning approach designed for hybrid air-underwater environments, and a comparison of hybrid to non-hybrid vehicle paths through simulated cave systems.} Another noteworthy feature of this work is grid-complete motion planning via a variation of PRM we call ``PRM on the go'' and a simple graph cost modification.

\subsection{Related Work} \label{related}
Related work has explored the challenges of dynamic re-planning with a high degree-of-freedom (DOF) systems, a high fill tunnel-like obstacle space, and a kinematic vehicle. Some methods use voxels, the 3D equivalent of 2D pixels, to discretize a 3D environment \cite{voxels}. Voxel representations are both practical and widely used, especially with compression algorithms like the OctoMap \cite{octomap}. Quaternion quadcopter control strategies have yielded successful positional \cite{Novel,umd} and attitude controllers \cite{attitude_quat,Bong}. Other approaches combine path planning with positional and attitude control to find optimal trajectories subject to physical dynamics \cite{opt_quad_traj,time_optimal}. 

To the authors knowledge, no studies have yet explored path planning for an air-underwater hybrid vehicle. Hybrid vehicle designs with some autonomous functionality have been presented \cite{slutgurs}, and other types of hybrid vehicles have been studied in the past, including: air-ground hybrid vehicles \cite{reviewer_cardrone} and VTOL-fixed wing vehicles 
\cite{reviewer_modeldrone}. Path planning for cooperative air and ground vehicles has also been studied \cite{coop-hybrid}. Our work differs from previous work in that we formulate and solve the energy minimization path planning problem for a single air-underwater hybrid vehicle.

\subsection{System Overview} \label{s:system_overview}
We use a layered positional/attitude controller to solve the two-point boundary value problem in both the air and underwater mediums. These solutions are then used within a dynamic sampling-based motion planning algorithm. Prior knowledge of the cave system is assumed to be either partial, inaccurate, or unknown. The vehicle has a limited battery life, which motivates a minimum energy solution. 

Our motion planning approach uses both a preprocessing and an online phase. A PRM motion graph and an initial path are found during preprocessing. Online, as the vehicle moves, the map is continually updated using on-board sensors, and the path is replanned accordingly (using D*-lite). 

To minimize battery use, the graph cost function is the expected energy consumed to travel across an edge. The air and water vehicle dynamics are used in a layered controller formulation shown in Fig. \ref{fig:layers}a, which is used to both inform the path planning algorithm as well as execute the path.

Most of the controller layers are structurally universal for air or underwater operation, only modifying some variable values. However, the positional controller works with different assumptions for both air and underwater.

The controller, motion planner, experiments, results, and conclusions are described in Sections \ref{s:control}, \ref{s:pathplanning}, \ref{s:experiment}, \ref{s:results}, and \ref{s:conclusion}.

\section{Controller Formulation} \label{s:control}

\begin{figure}[b!]
	\centering
	\includegraphics[height=5cm, trim=0 20 0 5, clip=true]{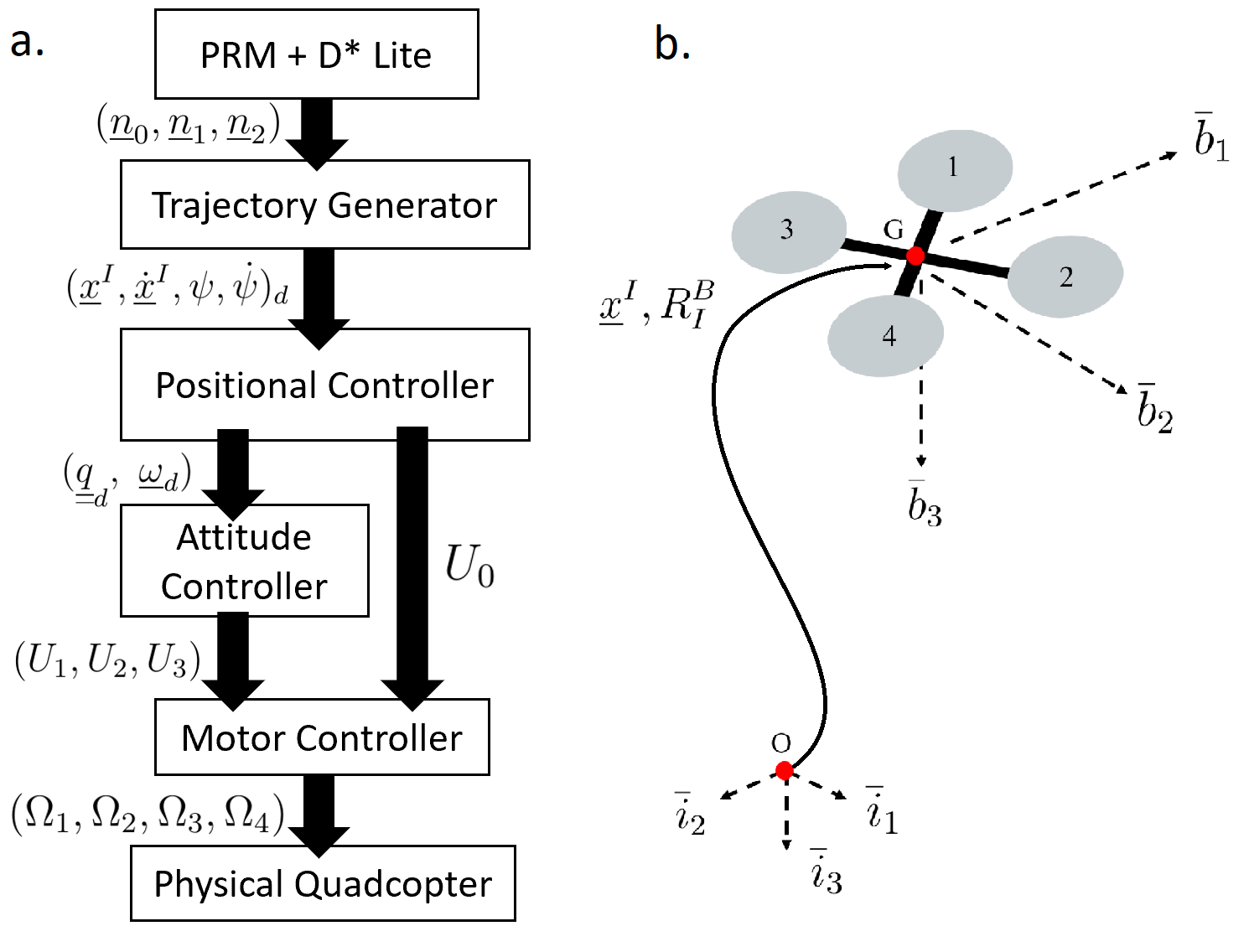}
	\caption{\textit{a. Layered path planning and control strategy used on the AQWUA. b. Rigid quadcopter and reference frames. }}
	\label{fig:layers}
\end{figure}

The overall system uses a coupled layered architecture consisting of a motion planner, trajectory generator, positional controller, attitude controller, and a motor controller. Each layer receives inputs from higher layers (Fig. \ref{fig:layers}a). 

The vehicle model is a standard X-configuration quad-rotor. However, when operating underwater the buoyancy decreases the weight vector, the body drag increases dramatically, and both rotor torque and thrust increase resulting in lower rotor-motor speed (affecting motor efficiency). 

The trajectory generator creates smooth functions of positions, velocities, and accelerations between nodes for the vehicle to track with the positional controller (Fig. \ref{fig:controller_kong}). Tracking is achieved by finding the necessary attitudes and attitude rates. The required moments, or torques, that the motors must generate about the vehicle's center of gravity are found by the attitude controller. Finally, the energy determination function finds the resulting rotor speeds and integrates electrical power to find the energy consumed based on experimental motor data.

\subsection{Trajectory Generator for Air and Water} \label{s:spline}

\begin{figure}[t!]
	\centering
	\includegraphics[height=2.7cm, trim=0 28 0 20, clip=true]{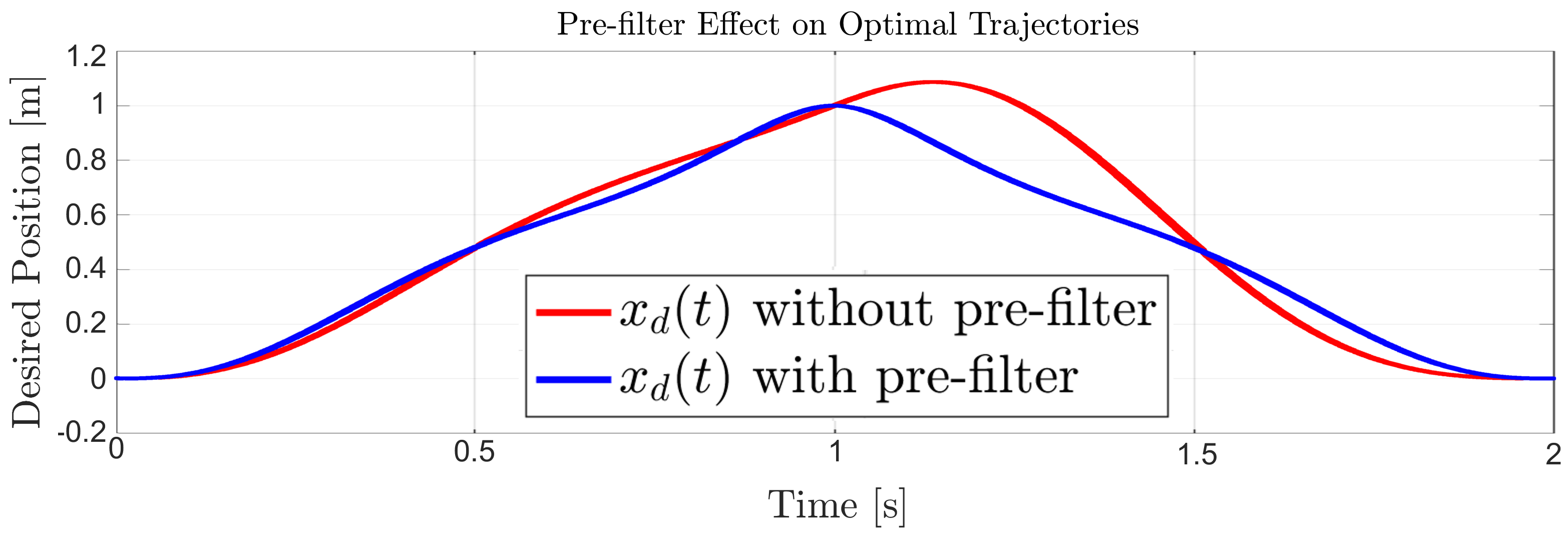}
    \caption{\textit{Resulting optimal trajectories with and without pre-filter algorithm. 1D case, desired x: $0 \to 1 \to 0$.}}
	\label{fig:prefilter}
\end{figure}

To smoothly navigate between path nodes $n_i$ and $n_{i+1}$ a polynomial trajectory $s_{i(i+1)}$ is used that relies on initial conditions at $n_i$ and the positions of $n_i,n_{i+1},$ and $n_{i+2}$. A cruise speed $v_c$ is chosen as the average velocity along the entire path. Consider the formulation of $s_{01}$ and $s_{12}$.
%
%
%
The arrival times $t_1$ and $t_2$ are found given an initial time $t_0$ and assuming a constant speed $v_c$ along straight line motion to each node. The vector velocity $\dot{\underline{x}}^I(t_i)$ at a node is defined with a magnitude of $v_c$ in the direction of the vector from $n_{i-1}$ to $n_i$, where an underline indicates a vector. This simple formulation can result in overshoot, which we minimize by pre-filtering (for a node, if the preceding and succeeding unit vectors have a component that changes sign, then that component of the node's velocity $\dot{\underline{x}}^I(t_i)$ is set to 0). The difference is evidenced in Fig. \ref{fig:prefilter} where an overshoot of 8.66\% is eliminated and the path length reduced by 6.73\%.

 
Two continuous, twice differentiable polynomials are required to spline the desired trajectory between nodes. Polynomial $\underline{s}_{ij}(t - t_i)$ goes from node $n_i$ to node $n_j$ from time $t_i$ to $t_j$.
For simplicity only the x coordinate polynomial $x_d(t)$ of $\underline{s}(t)$ is illustrated, as the method is the same in all directions. The following constraints are imposed:

\begin{enumerate}
  \item The position, velocity, and acceleration of node 0 at $t_0$ are:
  ${\underline{s}_{01}(t_0) = \underline{x}^I_0}$ and 
  ${\dot{\underline{s}}_{01}(t_0) = \dot{\underline{x}}^I_0}$ and 
  ${\ddot{\underline{s}}_{01}(t_0) = \underline{\ddot{x}}^I_0}$.
  \item The velocity, and acceleration of node 2 at $t_2$ are 0, so:
  ${\underline{s}_{12}(t_2) = \underline{x}^I_2}$ and
  ${\dot{\underline{s}}_{12}(t_2) = \underline{0}}$ and
  ${\ddot{\underline{s}}_{12}(t_2) = \underline{0}}$.
  \item Polynomials meet at $t_1$, so:
  ${\underline{s}_{01}(t_1) = \underline{s}_{12}(t_1) = \underline{x}^I_1}$ and
  ${\underline{\dot{s}}_{01}(t_1) = \underline{\dot{s}}_{12}(t_1) = \dot{\underline{x}}^I_1}$ and 
  ${\underline{\ddot{s}}_{01}(t_1) - \underline{\ddot{s}}_{12}(t_1) = \underline{0}}$
\end{enumerate}

To account for all the boundary conditions, a minimum of two $6^{th}$ order polynomials are required to describe $\underline{s}_{01}$ and $\underline{s}_{12}$. Such a formulation has been shown in other studies \cite{opt_quad_traj}. However, an explicit solution can be found by adding an extra term and minimizing the length of the trajectory. Let the polynomials be defined as:
${
 \underline{s}_{01}(t) = \sum_{i=0}^7 a_i(t-t_0)^i  \text{ for $t_0 \leq t < t_1$}
}$
and
${
  \underline{s}_{12}(t) = \sum_{i=0}^7 b_i(t-t_1)^i  \text{ for $t_1 \leq t \leq t_2$}
}$.

The cost function $J$ for this minimization problem is the total arc length of both polynomials:
$$\textstyle
    J(\underline{a}, \underline{b}) = \int_{t_0}^{t_1}\sqrt{1 + (\dot{\underline{x}}^I_{d,01}(t))^2} dt + \int_{t_1}^{t_2}\sqrt{1 + (\dot{\underline{x}}^I_{d,12}(t))^2} dt
$$

The optimization problem is to minimize $J(\underline{a}, \underline{b})$  subject to the constraints listed above. It can be explicitly solved for x, y, and h each time the vehicle arrives at a new node. We note that an acceleration based cost function would more accurately optimize for energy use but does not have an explicit solution. We consider three nodes, rather than two, because a smooth trajectory is ensured by the boundary conditions described above.

A positional controller that follows the trajectory requires different formulations for air and water. These are now described in section \ref{s:PosAir} and \ref{s:PosH20}, respectively.

\subsection{Positional Controller for Air} \label{s:PosAir}

The positional controller calculates desired attitudes $\doubleunderline{q}_d$ and angular velocities $\underline{\omega}_d$ to follow the positions, velocities, and accelerations given by the trajectory generator described above. This drives the vehicle's current state to the desired state by rotating its thrust vector. In air, the vehicle dynamics are: 

\begin{equation}
    m\underline{\ddot{x}}^I = R^B_I\underline{T}^B + m\underline{g}^I
    \label{eq:N2Lair}
\end{equation}
where $\underline{g}^I = [0 \ 0 \ 9.81]^T \ m/s^2$, any quantity superscript $I$ or $B$ indicates it is defined in the inertial frame $I$ or body frame $B$, $\underline{T}^B$ is the thrust vector, and $R^B_I$ describes the rotation from the body frame $B$ to the inertial frame $I$ depicted in Fig. \ref{fig:layers}b. 
We use a modified PD controller with gains $K_p$ and $K_d$ from \cite{quatPD, Novel, umd}  to follow the desired trajectory:
\begin{equation}
    \ddot{\underline{x}}^I = \underline{\ddot{x}}_d^I + K_p(\underline{x}^I_d - \underline{x}^I) + K_d(\underline{\dot{x}}_d^I - \underline{\dot{x}}^I)
    \label{eq:posPD}
\end{equation}

We find an acceleration vector $\bar{F}^I$ for the attitude controller to track by considering the inertial force acting on the aircraft from (\ref{eq:N2Lair}). To determine the inertial vector which orientates the quadcopter in the desired direction of travel rearrange (\ref{eq:N2Lair}) and (\ref{eq:posPD}) to arrive at:
${
    R^B_I\underline{T}^B = m\underline{\ddot{x}}^I - m\underline{g}^I = m\underline{\ddot{r}}^I - m\underline{g}^I 
}$
and
\begin{equation}
    R^B_I\underline{T}^B \overset{\Delta}{=} \underline{F}^I = m(\underline{\ddot{r}}^I - \underline{g}^I).
    \label{eq:_Uo}
\end{equation}

$\bar{F}^I$ is the unit vector of $\underline{F}^I$: the desired thrust vector in the inertial frame. In the body frame the unit thrust vector is $\bar{T}^B = [0 \ 0 \ -1]^T$. To make $\bar{F}^I$ and $\bar{F}^B$ co-linear, a rotation is required:
\begin{equation}
    \bar{F}^I = R^B_I\bar{T}^B = \doubleunderline{\hat{q}}^*_d \otimes 
    \begin{bmatrix}
    0 \\
    \bar{T}^B
    \end{bmatrix}
     \otimes \doubleunderline{\hat{q}}_d
    \label{eq:rotate}
\end{equation}
Where $\doubleunderline{\hat{q}}_d$ is the desired pitch and roll quaternion, $(\otimes)$ is the quaternion cross-product, (*) is the conjugate quaternion \cite{umd}. Using (\ref{eq:rotate}), the roll and pitch portions of the desired quaternion vector \cite{quat_diff} can be calculated using:
$$
    \doubleunderline{\hat{q}}_d = \frac{1}{\sqrt{2(1+\bar{F}^{B^T}\bar{F}^I)}}
    \begin{bmatrix}
    1+\bar{F}^{B^T}\bar{F}^I\\
    \widetilde{\bar{F}^B} \bar{F}^I \\
    \end{bmatrix}
$$
Where the tilde operator is a skew symmetric matrix, such that if $\underline{\omega} = [p \ q \ r]^T$, then $\widetilde{\omega}$ is described by:
$$
	\widetilde{\omega} =
	\begin{bmatrix}
	0 & -r & q \\
	r & 0 & -p \\
	-q & p & 0 \\
	\end{bmatrix}
$$

Only roll and pitch are encoded by the attitude vector $\doubleunderline{\hat{q}}_d$. A yaw angle correction is applied to find the desired quaternion vector $\doubleunderline{q}_d$ as follows:
${
    \doubleunderline{q}_d = \doubleunderline{\hat{q}}_d \otimes [\cos(\psi_d/2) \ 0 \ 0 \  \sin(\psi_d/2)]^T
}$.

To accurately track a moving reference signal $\underline{x}_d(t)$, the desired angular velocity vector $\underline{\omega}_d(t)$ must also be tracked \cite{omega_desired}, which is found by taking a time derivative of the thrust vector and applying the transport theorem:
\begin{equation}
    \frac{d}{dt}(\bar{T}^B) = \dot{\bar{T}}^B = \cancelto{0}{^B \frac{d}{dt}(\bar{T}^B)} + \widetilde{\omega}_d\bar{T}^B
    \label{eq:dF}
\end{equation}
$^B \frac{d}{dt}(\bar{T}^B) = 0$ because the thrust vector is fixed in $B$ and the derivative is taken with respect to $B$. Equation (\ref{eq:dF}) is rotated into the inertial frame \cite{umd, omega_desired, quatPD}, and the resulting desired angular velocities are 
${
    \underline{\omega}_d = \widetilde{\bar{F}^I} \dot{\bar{F}}^I
}$.
As before, this only defines the roll and pitch angular velocities. The desired yaw rate $\dot{\psi}_d$ is based on the path planning algorithm: $r_d = \dot{\psi}_d$. In this work $\psi = \dot{\psi} =0 $, as adjusting $\psi$ is energy inefficient due to the lack of control authority in yaw. The time derivative of the inertial thrust unit vector $\dot{\bar{F}}^I$ completes the formulation of $\underline{\omega}_d$: 
\begin{equation}
    \dot{\bar{F}}^I = \frac{\underline{\dot{F}}^I}{\norm{\underline{F}^I}} - \frac{\underline{F}^I(\underline{F}^{I^T}\underline{\dot{F}}^I)}{\norm{\underline{F}^I}^3}
    \label{eq:dFi}
\end{equation}
where the value of $\underline{\dot{F}}^I$ is found via numerical differentiation. 

This completes the formulation of the desired attitudes $\doubleunderline{q}_d$ and angular velocities $\underline{\omega}_d$ in air. This formulation appears in \cite{ Novel, umd, quatPD, omega_desired}, and lays the foundation for the novel positional controller in water presented below.


\subsection{Positional Controller for Water} \label{s:PosH20}

The formulation of the positional controller in water follows the same outline as its aerial counterpart with modifications in the underlying equation of motion to involve drag and buoyancy. Drag is neglected in air, but is non-negligible in water due to water's larger density. Equation (\ref{eq:posPD}) is used again and the outputs are the same. 

The underlying equation of motion in water is:
\begin{equation*}
    m\underline{\ddot{x}}^I = R^B_I\underline{T}^B + m\underline{g}^I  - \underline{F}^I_{buoy} + \underline{F}^I_{D}.
\end{equation*}
Where the drag function $\underline{F}^I_{D}$ is proportional to the characteristic area and the translation speed in the body frame: $\underline{\dot{x}}^B = R^I_B \underline{\dot{x}}^I$. 
Drag, is given by
${
    F^B_{D} = -\frac{1}{2}\rho C_DA |\underline{\dot{x}}^B|^T *\underline{\dot{x}}^B
}$ in the body frame \cite{Nelson}.
Flat plate areas for each cardinal direction, listed in table \ref{table:1}, are stored in the diagonal matrix $C_DA = diag([f_1 \ f_2 \ f_3])$. This quantity is rotated into the inertial frame via $\underline{F}^I_{D} = R^B_I \underline{F}^B_{D}$.

The effect of buoyancy can be viewed as a reduced factor of gravity $0 \leq b \leq 1$ such that
${
    m\underline{g}^I-  \underline{F}^I_{buoy} = bm\underline{g}^I
}$.
Here, $b = 1$ indicates no buoyancy where the motors must support the entire weight of the AQWUA, and $b = 0$ indicates the vehicle is totally buoyant. A buoyancy factor of 0.75 is used to ensure the motors do not saturate or stop spinning during nominal operation. The inertial acceleration vector $\bar{F}^I$ is:
${
    R^B_I\underline{T}^B \overset{\Delta}{=} \underline{F}^I = m\underline{\ddot{x}}^I - bm\underline{g}^I  - F^I_D = m\underline{\ddot{r}}^I - m\underline{g}^I - F^I_D  
}$
which simplifies to
${
    \underline{F}^I = m(\underline{\ddot{r}}^I - b\underline{g}^I - \frac{F^I_D}{m})
}$.

With these changes, determining the desired quaternion attitude $\doubleunderline{q}_d$ and angular velocity $\underline{\omega}_d$ follows the same pattern as the air case using Equations (\ref{eq:_Uo}) to (\ref{eq:dFi}). 

\subsection{Quaternion Attitude Controller for Air and Water}

Moments required to achieve the desired attitudes $\doubleunderline{q}_d$ and angular velocities $\underline{\omega}_d$ are found by the attitude controller. Because the effects of buoyancy and drag are accounted for in the positional controller the formulation of the attitude controller is the same in air and water. 

The AQWUA is assumed to be rigid and has a mass-moment of inertia matrix that is diagonal, such that $J = diag([J_x$ $J_y$ $J_z])$. Angular velocities around each axis, expressed in the body frame, are denoted as: $\underline{\omega} = [p \ q \ r]^T$. Euler’s second law \cite{umd,Nelson,compactQuat} describes the rate of change of the angular velocities in the body frame: 
${
	J\dot{\underline{\omega}} = \underline{U}_{1,2,3} -  \widetilde{\omega}J\underline{\omega} - D_w\underline{\omega}
}$
$\underline{U}_{1,2,3}$ = $[U_1$ $U_2$ $U_3]^T$ is the attitude control vector, or the body moments generated around each axis by the motors, and the matrix $D_w$ is the attitude drag matrix, which changes between air and underwater.

The rate of change of a quaternion \cite{umd} is related to $\underline{\omega}$:
\begin{equation}
	\dot{\doubleunderline{q}}(t) =
	\begin{bmatrix}
	-\frac{1}{2}\underline{q}^T\underline{\omega} \\
	\frac{1}{2}(\widetilde{\underline{q}}+I_{3,3}q_{0})\underline{\omega} \\
	\end{bmatrix}
	\label{eq:quatEOM}
\end{equation}
Where the double underline represents the entire quaternion, the single underline represents the vector quaternion and $I_{3,3}$ is the identity matrix.
The quaternion tracking error is:
\resizebox{.48\textwidth}{!}{%
$
	\doubleunderline{q}_{e} = 
	\begin{bmatrix}
	q_{0e} \\
	q_{1e} \\
	q_{2e} \\
	q_{3e} \\
	\end{bmatrix} =
	\begin{bmatrix}
	q_{0d} & q_{1d} & q_{2d} & q_{3d} \\
	-q_{1d} & q_{0d} & q_{3d} & -q_{2d} \\
	-q_{2d} & -q_{3d} & q_{0d} & q_{1d} \\
	-q_{3d} & q_{2d} & -q_{1d} & q_{0d} \\
	\end{bmatrix}
	\begin{bmatrix}
	q_{0m} \\
	q_{1m} \\
	q_{2m} \\
	q_{3m} \\		
	\end{bmatrix}
	= \doubleunderline{q}^*_{d} \otimes \doubleunderline{q}_{m}
$}
where $d$ and $m$ denote the desired and measured quantities, respectively. If two bodies have the same attitude then ${\doubleunderline{q_e}=[1\,\, 0\,\, 0\,\, 0]^T}$. 
The attitude controller 
is:
\begin{equation}
	\underline{U}_{1,2,3} = -K_p\underline{q}_e - K_d\underline{\omega}_e 
	\label{eq:C1} 
\end{equation}
where $K_p$ and $K_d$ are positive definite diagonal gain matrices and vary based on medium. It was found by \cite{Bong, Novel, Khalil} that this controller is asymptotically stable, by use of both Lyanpunov and LaSalle analysis. 

The error state equations may be written \cite{compactQuat}:

\begin{equation}
	J\dot{\underline{\omega}}_e = \underline{U}_{1,2,3} -  \widetilde{\omega}_e J\underline{\omega}_e
	\label{eq:compactE2L} 
\end{equation}
\begin{equation*}
	\dot{\doubleunderline{q}}_e = 
	\begin{bmatrix}
	-\frac{1}{2}\underline{q}_e^T\underline{\omega}_e \\
	\frac{1}{2}(\widetilde{\underline{q}}_e+I_{3,3}q_{0e})\underline{\omega}_e \\
	\end{bmatrix}
\end{equation*}
\begin{equation*}
	 \norm{\doubleunderline{q}}_e = 1 = q_{0e}^2 + q_{1e}^2 + q_{2e}^2 + q_{3e}^2
\end{equation*}
and the positive definite Lyapunov function candidate and its derivative are then:
\begin{equation*}
    V = \frac{1}{2}\underline{\omega}_e^T K_p^{-1} J \underline{\omega}_e + (q_{0e}-1)^2 + q_{1e}^2 + q_{2e}^2 + q_{3e}^2 
\end{equation*}
\begin{equation}
    \dot{V} = -\underline{\omega}_e^T K_p^{-1}K_d\underline{\omega}_e 
    \label{eq:mVdef}
\end{equation}

As long as attitude gains $K_p$ and $K_d$ are positive definite, $\dot{V}$ is negative definite for $\omega_e$. Equation (\ref{eq:mVdef}) can be used in a LaSalle analysis \cite{Khalil} with the invariance condition of $\underline{\omega}_e = \underline{\dot{\omega}}_e = \underline{0}$ for all time. Combining (\ref{eq:compactE2L}) and (\ref{eq:C1}) and substituting the invariance condition shows that $\underline{q}_{e}$ goes to 0, proving asymptotic stability.

\subsection{Energy Determination for Air and Water} \label{s:energydetermin}

A Simulink program is developed to test the layered control architecture and record vehicle energy usage. In addition to characterizing the execution of a path, the simulation is used to find the stop-stop energy consumption between two nodes to inform edge costs. Stop-stop energy consumption refers to the energy used by the motors to move the vehicle along an edge where the initial and final conditions are both at rest. A model for the motor controller as depicted in \mbox{Fig.\ \ref{fig:layers}a} is used to estimate the electrical power consumption and is described below.

Rotor thrust and torque are defined: $T_i = C_T\rho A (\Omega_i R)^2 = K_T \Omega_i^2$ and $Q_i = C_Q\rho A (\Omega_i R)^2 R = K_Q \Omega_i^2$ where $C_T$ and $C_Q$ are the rotor thrust and torque coefficients respectively, $\rho$ is air density, $\Omega$ is rotor speed, $A$ is rotor disk area, and $i$ indicates an association to rotor $i \in [1, 2, 3, 4]$, as described in \cite{leishman2006principles}. The control vector $\underline{U} = [U_0 \ \underline{U}_{1,2,3}^T]^T$ is mapped to motor RPM via standard X configuration.

An example of overall performance of the entire controller stack is illustrated in Fig.\ \ref{fig:controller_kong}, which appears in the appendix.

\begin{table}[t!]
\centering
\caption{Parameters of AQWUA vehicle.}
\label{table:1}
\begin{tabular}{|c | c | c | c|} 
 \hline
Parameter & Value & Parameter & Value\\\hline
$J_{xx}$ & 0.0165 kg-m$^2$ & $C_T$ & 0.0103 \\
$J_{yy}$ & 0.0324 kg-m$^2$ & $C_Q$ & 0.00118 \\
$J_{zz}$ & 0.0385 kg-m$^2$ & $f_{1,2}$ & 0.01 \\
$m$  & 3.865 kg & $f_3$ & 0.03 \\
$L$  & 200 mm & $R$ & 190.5 mm \\
\hline
\end{tabular}
\end{table}

The relationship between motor RPM and DC electrical power was determined experimentally in air and water. The electrical power for each motor along a trajectory is found and summed to find $P_{Total}$. The energy used is:
${
    E(t_s) = \int_{0}^{t_s} P_{Total}(t) dt
}$.

A lookup table of stop-stop energy costs for each integer displacement vector relative within a 16x16x16 meter cube centered at the origin is found for the air and underwater cases using the controllers described in Section \ref{s:control}, this informs the path planning edge costs. A stop-stop maneuver is considered complete when the 2\% settling criterion is reached in each of the X,Y, and H directions.

\section{Path Planning for Air and Underwater} \label{s:pathplanning}
To navigate a dynamic cluttered environment with both air and water the path planner uses the process laid out below:

\begin{enumerate}
   \item[1.] Preprocessing
   \begin{itemize}
     \item PRM creates graph with N nodes connected to all nodes within some radius.
     \item The trajectory look-up table estimates the cost for each edge as stop-stop consumed energy.
   \end{itemize}
   \item[2.] Online (these actions occur in a loop)
   \begin{itemize}
       \item Update the graph for all differences between map and environment within sensor visibility.
       \item Add new random nodes to the graph within the sensor visibility (PRM on the go).
       \item Use D*-Lite to update the graph edge costs and current path to goal.
       \item Pass the next several nodes to the trajectory calculator to obtain control inputs.
       \item Use control inputs to move to the next node.
       \item If no path is found, follow edge case procedure.
       \end{itemize}
\end{enumerate}

Implementing a cost modification to D*-Lite, PRM on the go, and an edge case procedure allows for a grid-complete path planning algorithm, see section \ref{s:HiRes_PP} for details. 

\subsection{Overview} \label{s:environ}
The workspace $W$ that this problem resides in is a $\rm I\!R^3$ occupancy grid, where obstacles $O \in W$ are represented as a boolean: present, or not present. 

A node $n$ represents one voxel and the vehicle is represented as a single voxel. All obstacle edges are extended by one vehicle radius to justify this representation regardless of resolution. This presents a collision rule: any line that intersects a vertex or edge is said to intersect all voxels sharing this vertex or edge. For example, if each voxel is centered at integer X, Y, H coordinates with a side length of 1, then a line connecting (0,0,0) to (1,0,0) would intersect 2 voxels, to (1,1,0) would intersect 4 voxels, to (1,1,1) would intersect 8 voxels.

Not all obstacles will be known before the AQWUA begins exploring a cave, which motivates a dynamically modeled environment. The vehicle stores a map $M$ of the workspace with an assumed distribution of obstacles $\bar{O} \in M$. As the vehicle moves, sensor readings confirm the actual state of voxels in the environment and updates the stored map. 

\subsection{Pre-Processing} \label{s:preprocessing}

In order to generate a graph $G$ {\it a priori} a modified probabilistic roadmap (PRM) approach is used. Let $N$ denote the number of unique nodes $n_1,n_2,...,n_N$ that are randomly sampled. All nodes within a radius $r_{MaxEdge}$ are connected. The air-to-underwater and underwater-to-air transition edges are limited to the vertical case (because hybrid multi-rotor craft have yet to achieve a non-vertical transitions).

A modification of the algorithm in \cite{3dvoxelray} for calculating voxel-line intersection is used to collision check edges. It returns the coordinates of all voxels a line intersects in accordance with the collision rule in section \ref{s:environ}.
%
Edge costs not in collision with any obstacle $\bar{O} \in M$ are determined via lookup from the cost table described in section \ref{s:energydetermin}.

A false ``no path exists" conclusion issue may arise when the initial map indicates only one path to the goal, but is found to be obstructed by sensor readings. In this event the traditional D* lite algorithm falsely determines that no path exists while there are obstacles present in the assumed map that do not exist in the workspace. To address this issue, the following a cost modification is introduced: all nodes and edges that are found to be in collision with an assumed obstacle $\bar{O}$ are assigned a large finite cost $C_{large}$ rather than discarding the node or edge. $C_{large}$ is a value greater than the cost to traverse the entire environment twice in its largest dimension. In a standard implementation of PRM, any edge that intersects any obstacle is declared to have an infinite cost and disregarded. In our modification, an infinite cost is only assigned to edges that are confirmed to be in collision with obstacles based on sensor readings $O$ gathered at runtime.

\begin{figure}[t!]
	\centering
	\includegraphics[width=.5\textwidth, trim=0 0 0 0, clip=true]{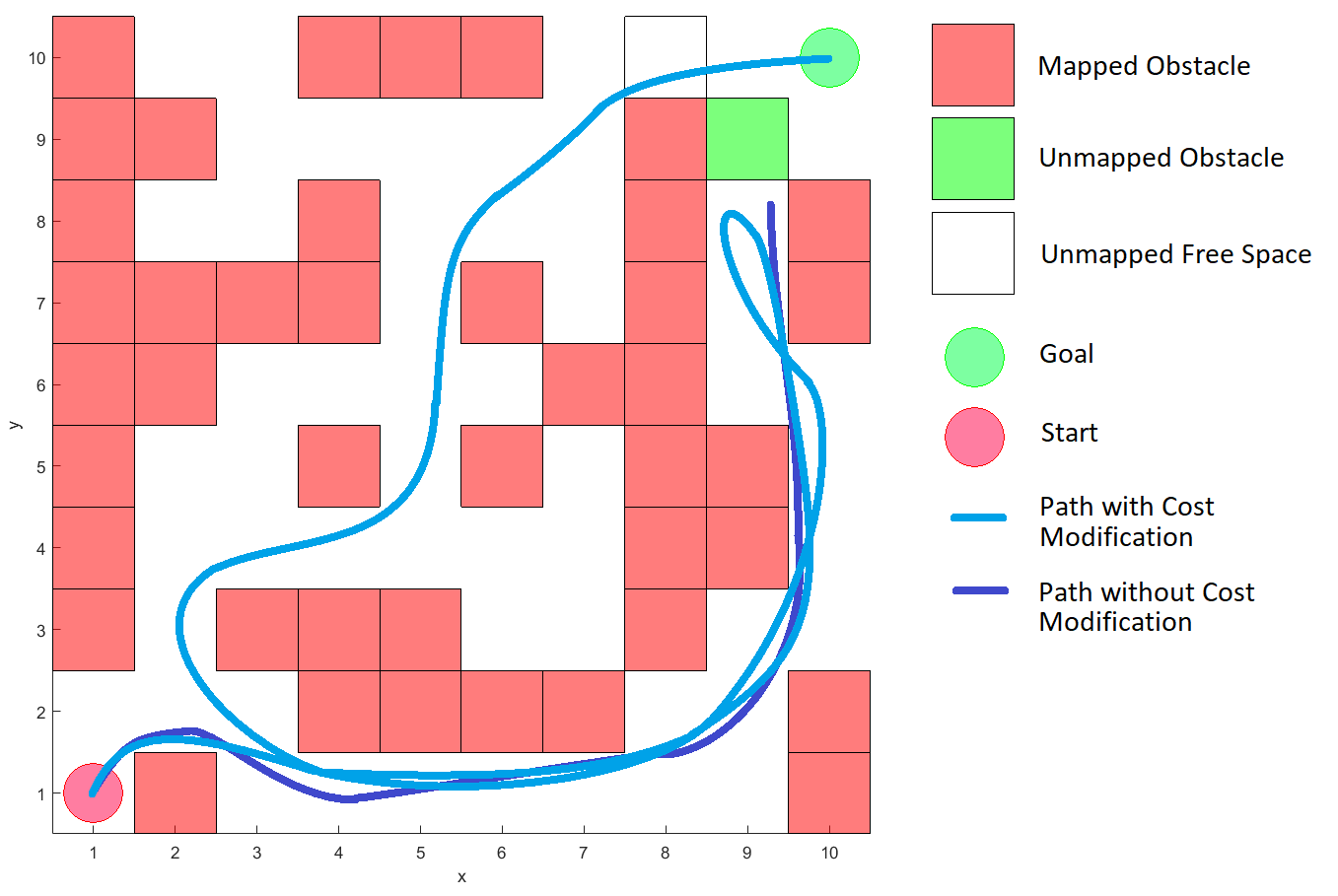}
	\caption{\textit{ Cartoon depicting how the Cost modified D*-Lite algorithm solves the problem while the standard version does not.}}
	\label{fig:26con_case}
\end{figure}

The cost modification effectively forces the vehicle to explore all initially known free pathways to the goal, then explore pathways to the goal through unconfirmed obstacles. Only when every path to the goal is blocked by sensor confirmed obstacles does the vehicle correctly conclude that no path exists.
A simple 2D example is presented in Fig. \ref{fig:26con_case}. The D*-lite implementation without the cost modification is unable to solve for a path whereas the implementation with the modification is able to.

It is also important to note that the stop-stop cost is used as an approximation for the energy consumed during a smooth execution of an edge. Such a dynamically executed edge is dependent on the initial conditions and therefore node history as described in section \ref{s:spline}. As a result, it becomes impossible to assign a dynamically executed cost to an edge in preprocessing.

\subsection{Planning} \label{s:HiRes_PP}
In the online stage, a modified D*-Lite is used to solve the graph generated in preprocessing. The modifications include the addition of new nodes to the graph via PRM on the go and the definition of an edge case procedure.

\begin{figure}[h!]
	\includegraphics[height=3.5cm, trim=0 05 0 5, clip=true]{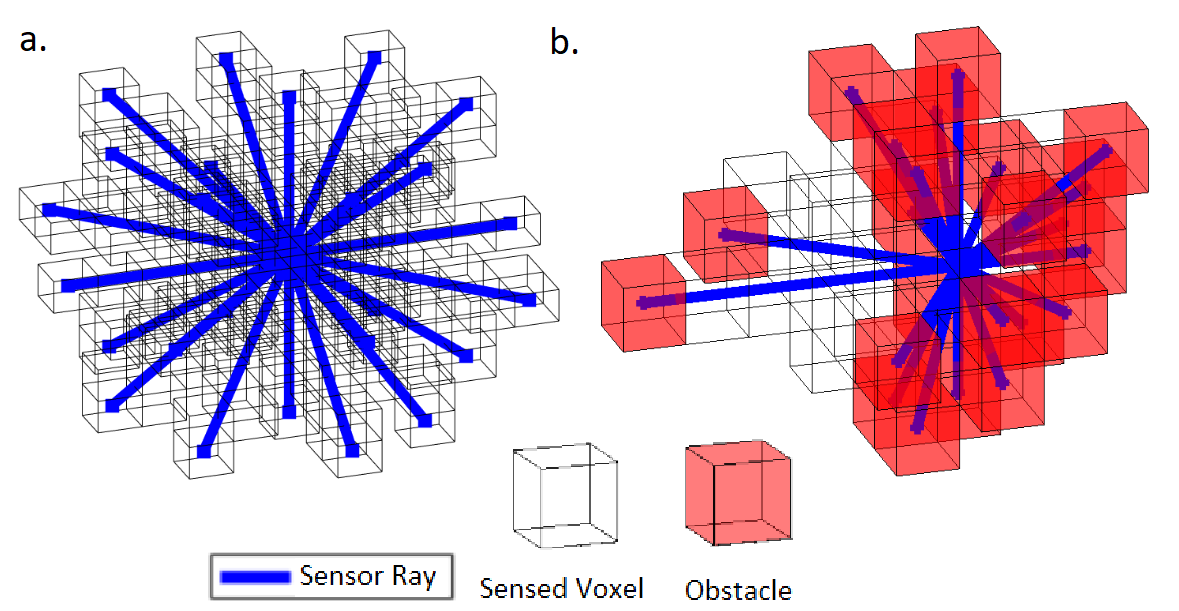}

	\caption{\textit{a. All voxels sensed by a 45 degree resolution sensor array with radius 5 in a free environment. b. All voxels checked by same array in an environment with obstacles, shows lack of information behind an obstacle.}}
	\label{fig:sensors}
\end{figure}

To model an onboard sensor array the vehicle sends out multiple evenly radially distributed rays of a given length using the voxel-line intersection function described in section \ref{s:preprocessing}. In Fig. \ref{fig:sensors}a the voxels read by a sensor array with 45 degree resolution and radius of 5m in an empty environment are shown. The decreasing resolution of readings with distance can be seen. Additionally the sensor readings do not return data behind obstacles, illustrated in Fig. \ref{fig:sensors}b. The sensor implementation also does not return data across the water surface as is the case with actual laser or sonar systems. This sensor model is used to check the environment against the map and account for any differences. The set of voxels checked by the sensors at node $n_i$ is $S_i \subset W$ and any obstacles found are $O \in W$ prompting an infinite cost for edges in collision.

To allow for a theoretical probabilistic grid completeness guarantee a process called ``PRM on the go" is introduced. PRM on the go randomly chooses a predetermined amount of its sample points from within the set of voxels checked by sensors $S_i \subset W$ and adds new nodes according to the standard PRM process.

The edge case procedure is dependent on the mission parameters. Practically, if a vehicle found no path in the online stage it would immediately conclude that no path exists and route the vehicle back to the start node. 
However, it is possible that a path is not found due to insufficient node distribution. For the purposes of algorithmic completeness in a discrete space, we modify the edge case procedure follows: for a graph $G$, if every path $P_o \subset G$ has cost $C_o = \infty$ then the vehicle adds nodes $n_i \subset S_{all}$ to $G$ until either $\exists P_o$ with cost $C_o < \infty$ or $\forall p_i \in S_{all}, p_i \in G$, where $S_{all} \subset W$ is the union set of all $S_b$ where $\forall b, n_b \in G$ has a finite cost to start. This edge case behavior samples all possible voxels before determining no path exists; thus the algorithm is resolution complete \cite{completeness_proof} \cite{sampling_proofs} with respect to the voxel discritization.
PRM on the go is implemented in this work to allow for incrementally lower cost paths and prevention of false no path exists conclusions, the practical edge case procedure is implemented if no path is found.




\subsection{Integration with Trajectory Creation}

In order to generate a trajectory, at each movement step the path planning algorithm outputs the current node coordinates and conditions, and the next two node coordinates in the optimal path to goal. 
These coordinates, and conditions allow for a trajectory to be calculated as per section \ref{s:spline} from the first node, through the second, to the third. This trajectory is then followed to the second node and the process is repeated to ensure the vehicle does not stop moving and has favorable initial conditions if re-plan does not occur. If a re-plan does occur, then the third node from the original trajectory is no longer the second node in the new trajectory.
This strategy allows for the creation of a continuous, smooth, trajectory and subsequent control inputs for a given path to the goal, regardless of dynamic re-planning that may occur.

\section{Experimental Setup} \label{s:experiment}

To quantify the benefits of the hybrid path planner Monte Carlo simulations were conducted where air-only, water-only, and hybrid path planners attempted to find a path in many environments. 

200 procedurally generated submerged cave environments are created as per the procedure described in section \ref{s:appendix}. Two sets of start and goal nodes are quasi-randomly chosen for each environment, one set has both endpoints in air, the other in water. The endpoints are chosen to be free spaces within some margin of the x-axis boundaries. 
All path planners are identical, but operate on different graphs created by PRM. The non-hybrid path planners were restricted to nodes only in the air or water, the hybrid was not restricted. All graphs had equal node density.

\begin{figure}[t!]
	\centering
	\centering
	\includegraphics[height=5.5cm, trim=0 10 0 10, clip=true]{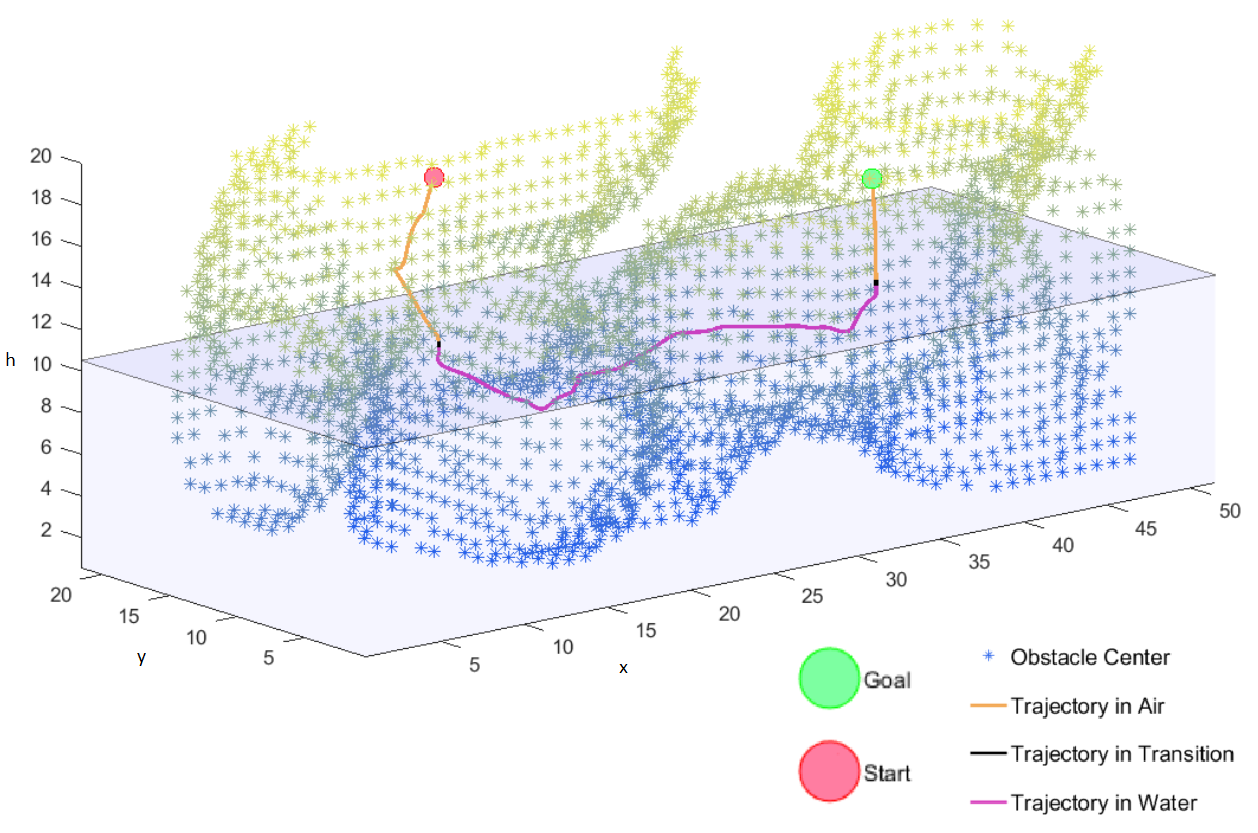}

	\caption{\textit{The trajectory a hybrid vehicle takes between two air-air endpoints}}
	\label{fig:cave_traject}
\end{figure}

The air-only and hybrid path planners attempt to find a path between the air-air endpoints whereas the water-only and hybrid path planners attempt to find a path between the water-water endpoints. An example of a hybrid trajectory through a cave is shown in Fig. \ref{fig:cave_traject}.

\section{Results} \label{s:results}
 Comparing the ability of the hybrid vehicle and non-hybrid vehicles to solve the same problems reveals the extent of the benefits that a multi-medium vehicle provides. Fig.\ \ref{fig:costs} summarizes the results. 
 
 Each environment presented two problems to solve (air and water), so of the 400 problems, the non-hybrid planners were able to solve 49.75\% and the hybrid planner solved 70.5\%. A $\chi^2$ analysis with a null hypothesis that hybrid and non-hybrid planners have an equal probability to solve a given problem results in a test statistic of $2E-9$. For a default confidence value of 95\% this test clearly supports that the non-hybrid path planners do not have the same probability to solve a given problem as the hybrid; as is expected.
 
 
 Of the 200 air problems, 104 were solved by both the hybrid and air-only planners. When comparing path costs of solutions to those 104 problems the hybrid planner is on average 27\% more efficient than the air-only planer. However, when comparing the 94 water problems that both the hybrid and water-only planner could solve the hybrid planner is only 2.7\% more efficient. Both the air and water problem sets show an equal variance in energy usage along the trajectory using an F test, and a T test reveals that the difference in average trajectory cost for air is statistically significant with a 95\% confidence value, whereas in water it is not.
 
 The graph (stop-stop) cost was able to predict the air-only trajectory cost within an absolute difference of 3.4\%, however that difference increases to 23.2\%, 20.4\%, and 20.8\% for the hybrid (air problem), hybrid (water problem), and water-only trajectories respectively. Clearly the involvement of the water controller worsens the prediction. However, several other factors also play a role. The average speed of the vehicle is larger than the stop-stop speed for 99\% of the trajectories, and the actual trajectory length is longer than the graph length for all trajectories. A linear regression of \% prediction absolute difference against the \% speed increase shows a positive slope for all cases. This is true for the regression against total path length as well, indicating that the longer the path, the faster it's execution, the worse the stop-stop energy costs predict the actual trajectory cost.
 
\begin{figure}[t!]
	\centering
	\includegraphics[height=5cm, trim=0 10 0 25, clip=true]{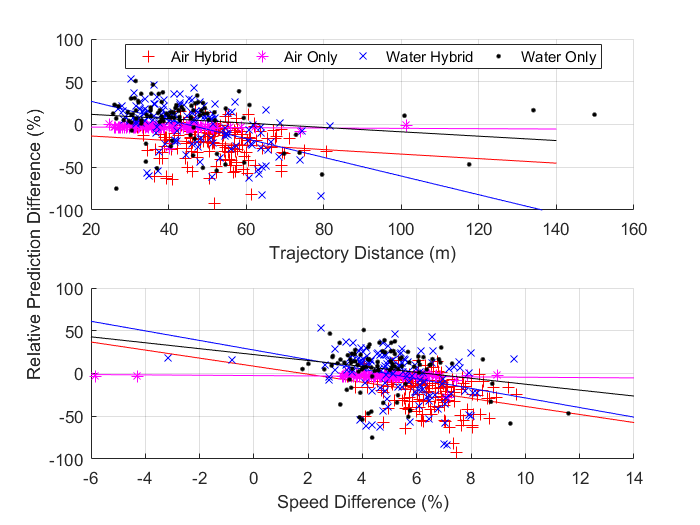}
	\caption{\textit{Top: \% relative cost prediction difference versus total trajectory length. Negative regression slope shows worse under-prediction of trajectory cost with increasing trajectory length. Bottom: \% relative cost prediction difference versus \% average trajectory speed increase. Negative regression slope shows worse under-prediction of trajectory cost with increasing average speed. }}
	\label{fig:precentdiff}
\end{figure}

 It was expected that the graph (stop-stop) energy cost would be higher than the executed trajectory energy cost, however this is not always the case. Of the 481 trajectories 65\% had a graph cost lower than the executed cost. To evaluate this, the relative \% difference of the graph prediction was found, a positive value indicates the graph cost over-predicted the trajectory cost which is expected. A linear regression of \% prediction relative difference against the \% speed increase and the total trajectory length shows a negative slope for all cases, see Fig. \ref{fig:precentdiff}. This supports that the longer the path, the faster it's execution, the more likely it is for the dynamic trajectory to have a cost higher than the stop-stop prediction.
 
 Modifying $v_c$ for air and water, and gains $K_p$, $K_d$ for the air positional controller, water positional controller, and attitude controller will all influence the energy usage to move between nodes. 

\def\HS{\hspace{\fontdimen2\font}}

\section{Conclusion} \label{s:conclusion}

\begin{figure}[t!]

 \HS \HS \HS \HS Energy Cost, mean and standard error over 400 trials

   \hspace{.6cm}
   \begin{minipage}{2.0cm}
    \small Air Scenerios 
   \end{minipage}
   \begin{minipage}{2.2cm}
   \small Water Scenerios 
   \end{minipage}
   \begin{minipage}{2.0cm}
   \small All Scenerios 
   \end{minipage}

\vspace{.1cm}

   \begin{minipage}{1.6cm}
     \begin{xy}
      \xyimport(100,100){
\includegraphics[height=1.8cm, trim=70 70 485 640, clip=true]{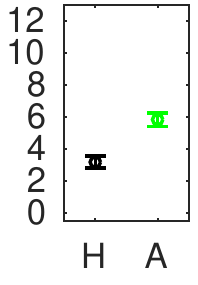}}
       ,(60,110)*{\text{\footnotesize Graph}}
       ,(-5,50)*{\rotatebox{90}{\text{\footnotesize  $10^5$ Joules}}}
     \end{xy}
     \end{minipage}
     \begin{minipage}{1.2cm}
     \begin{xy}
      \xyimport(100,100){
\includegraphics[height=1.8cm, trim=88 70 485 640, clip=true]{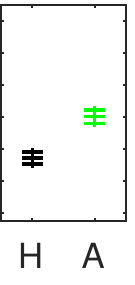}}
       ,(55,110)*{\text{\footnotesize Actual}}
     \end{xy}
     \end{minipage}
     \begin{minipage}{1.2cm}
     \begin{xy}
      \xyimport(100,100){
\includegraphics[height=1.8cm, trim=88 70 485 640, clip=true]{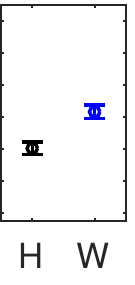}}
       ,(55,110)*{\text{\footnotesize Graph}}
     \end{xy}
     \end{minipage}
     \begin{minipage}{1.2cm}
     \begin{xy}
      \xyimport(100,100){
\includegraphics[height=1.8cm, trim=88 70 485 640, clip=true]{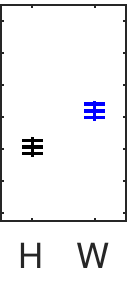}}
       ,(55,110)*{\text{\footnotesize Actual}}
     \end{xy}
     \end{minipage}
     \begin{minipage}{1.2cm}
     \begin{xy}
      \xyimport(100,100){
\includegraphics[height=1.8cm, trim=88 70 485 640, clip=true]{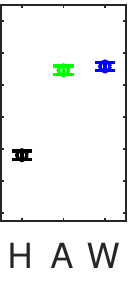}}
       ,(55,110)*{\text{\footnotesize Graph}}
     \end{xy}
     \end{minipage}
     \begin{minipage}{1.2cm}
     \begin{xy}
      \xyimport(100,100){
\includegraphics[height=1.8cm, trim=88 70 485 640, clip=true]{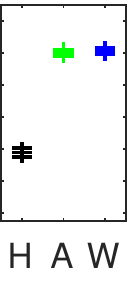}}
       ,(55,110)*{\text{\footnotesize Actual}}
     \end{xy}
     \end{minipage}
     \begin{minipage}{1.2cm}
     \begin{xy}
      \xyimport(100,100){
\includegraphics[height=1.8cm, trim=0 0 0 0, clip=true]{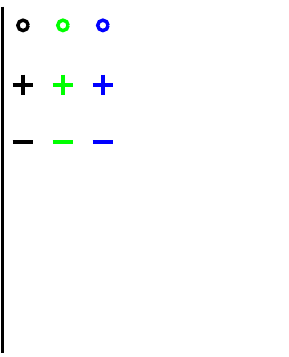}}
       ,(65,80)*[r]{\rotatebox{-35}{\text{\tiny Graph mean}}}
       ,(65,65)*[r]{\rotatebox{-35}{\text{\tiny Actual mean}}}
       ,(65,50)*[r]{\rotatebox{-35}{\text{\tiny std. error}}}
       ,(18,30)*{\rotatebox{90}{\text{\tiny Hybrid}}}
       ,(29,30)*{\rotatebox{90}{\text{\tiny Air Only}}}
       ,(40,30)*{\rotatebox{90}{\text{\tiny Water Only}}}
     \end{xy}
     \end{minipage}

     \begin{minipage}{7.5cm}
     \centering 

     {\small vehicle type}

     \end{minipage}

\vspace{.2cm}

\resizebox{\hsize}{!}{%
\begin{tabular}{|c | c | c || c | c |} 
 \hline
 & \multicolumn{2}{|c||}{Air Scenarios} & \multicolumn{2}{c|}{Water Scenarios}\\ \cline{2-5}
 & Hybrid & Air-Only & Hybrid & Water-Only\\ \hline
 Number Completed & 148 & 104 & 134 & 95\\
 Prediction $|\%|$ difference  & 23.17 & 3.38 &	20.42 &	20.81\\
\hline
\end{tabular}%
}

   \caption{\textit{Top: Mean energy used over 200 water-to-water and 200 air-to-air problem instances. Runs that fail are assumed to use all $1.2 \times 10^6$ Joule of stored energy (${\approx 20\%}$ more than that required by the longest run). `Graph' and `Actual' denote the graph cost (prediction) and the actual cost used by three types of vehicles (air-only, underwater-only, and hybrid air-underwater). Left 4 plots: Air and water problems are considered separately so that air and water vehicles are only evaluated in their native medium. Right 2 plots: All problems are combined such that air-only and water-only vehicles fail to solve problems in the other medium (exhausting batteries). Bottom: The number of runs completed in each type of scenario by each vehicle, and the \%-difference between graph and actual costs.}}
   \label{fig:costs}
\end{figure}


This work explores a novel water plus air motion planning problem faced by a hybrid submersible quadrotor vehicle in a submerged cluttered cave environment. The method we present to solve this problem  uses elements of both sample based and  graph search re-planning algorithms. It is found that the AQWUA's path is often more energy efficient than that of traditional air or submersible vehicles, and it is also able to solve a larger proportion of problems than air or submersible vehicles. 

We find that the graph energy use does not accurately predict the actual energy used, but the two quantities means are correlated across different problems.

The layered quaternion based kinematic controller is a key feature of this work. It adjusts based on operating medium, is asymptotically stable, and solves the two point boundary value problem with at rest initial and final conditions. It also finds a smooth trajectory through a series of nodes with arbitrary initial conditions, and can be scaled for greater resolution, or modified for greater complexity.  

\begin{figure*}[t!]

   \begin{minipage}{.24\textwidth}
   \begin{xy}
    \xyimport(100,100){\includegraphics[width=3.8cm, trim=75 75 340 630, clip=true]{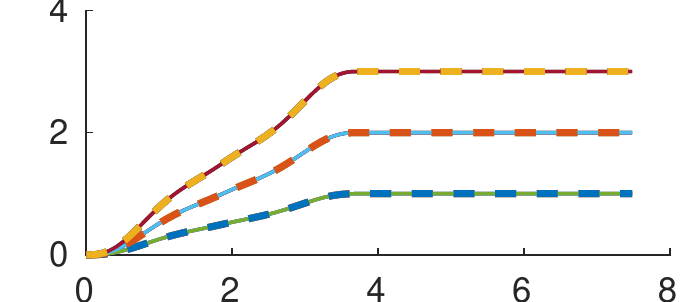}}
     ,(55,95)*{\text{Air Position}}
     ,(-5,50)*{\rotatebox{90}{\text{m}}}
     ,(80,25)*{\text{time (s)}}
   \end{xy}
   \end{minipage}
\hspace{.1cm}
   \begin{minipage}{.24\textwidth}
   \begin{xy}
    \xyimport(100,100){\includegraphics[width=3.8cm, trim=75 75 340 630, clip=true]{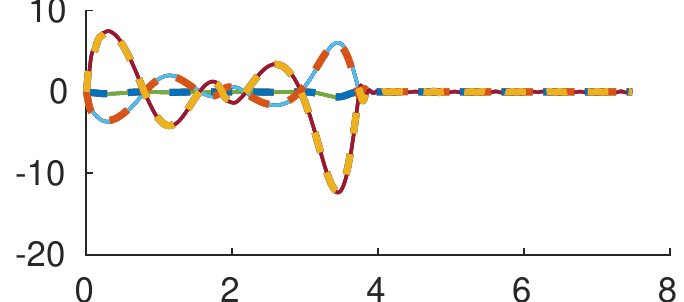}}
     ,(55,95)*{\text{Air Orientation}}
     ,(-5,50)*{\rotatebox{90}{\text{deg}}}
     ,(80,25)*{\text{time (s)}}
   \end{xy}
   \end{minipage}
   \hspace{.15cm}
   \begin{minipage}{.24\textwidth}
   \begin{xy}
    \xyimport(100,100){\includegraphics[width=3.8cm, trim=75 75 340 630, clip=true]{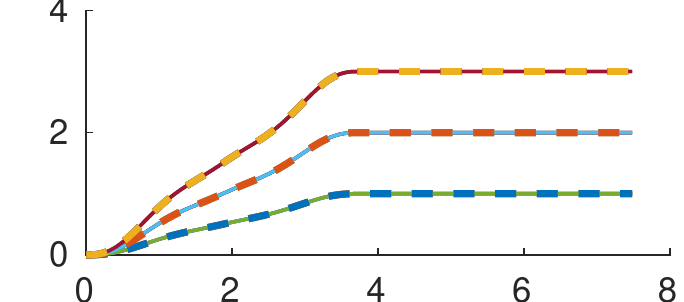}}
     ,(55,95)*{\text{Water Position}}
     ,(-5,50)*{\rotatebox{90}{\text{m}}}
     ,(80,25)*{\text{time (s)}}
   \end{xy}
   \end{minipage}
\hspace{.1cm}
   \begin{minipage}{.24\textwidth}
   \begin{xy}
    \xyimport(100,100){\includegraphics[width=3.8cm, trim=75 75 340 630, clip=true]{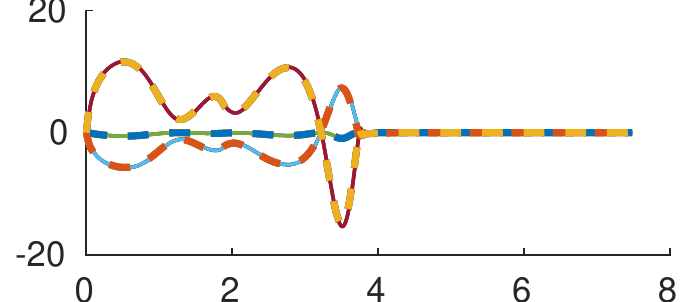}}
     ,(55,95)*{\text{Water Orientation}}
     ,(-5,50)*{\rotatebox{90}{\text{deg}}}
     ,(80,25)*{\text{time (s)}}
   \end{xy}
   \end{minipage}\\

\vspace{.15cm}

   \begin{minipage}{.24\textwidth}
   \begin{xy}
    \xyimport(100,100){\includegraphics[width=3.8cm, trim=75 75 340 630, clip=true]{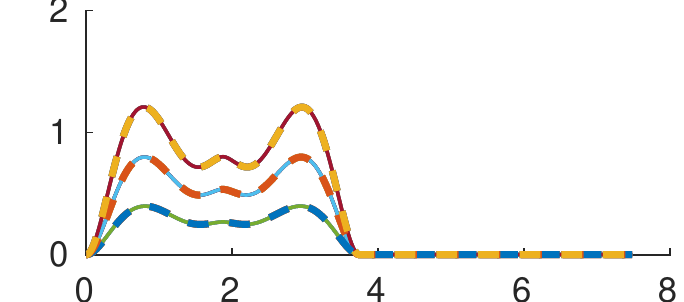}}
     ,(55,85)*{\text{Air Velocity}}
     ,(-5,50)*{\rotatebox{90}{\text{m/s}}}
     ,(80,25)*{\text{time (s)}}
   \end{xy}
   \end{minipage}
\hspace{.1cm}
   \begin{minipage}{.24\textwidth}
   \begin{xy}
    \xyimport(100,100){\includegraphics[width=3.8cm, trim=75 75 340 630, clip=true]{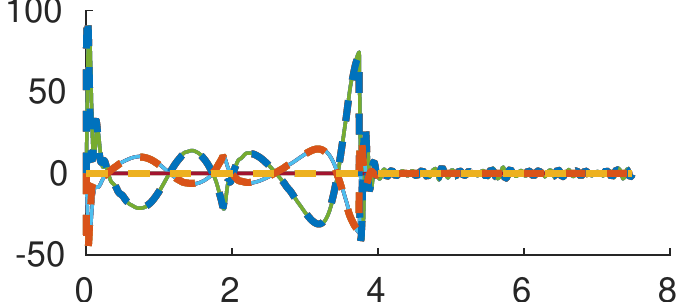}}
     ,(55,85)*{\text{Air Body Rates}}
     ,(-5,50)*{\rotatebox{90}{\text{deg/s}}}
     ,(80,25)*{\text{time (s)}}
   \end{xy}
   \end{minipage}
   \hspace{.15cm}
   \begin{minipage}{.24\textwidth}
   \begin{xy}
    \xyimport(100,100){\includegraphics[width=3.8cm, trim=75 75 340 630, clip=true]{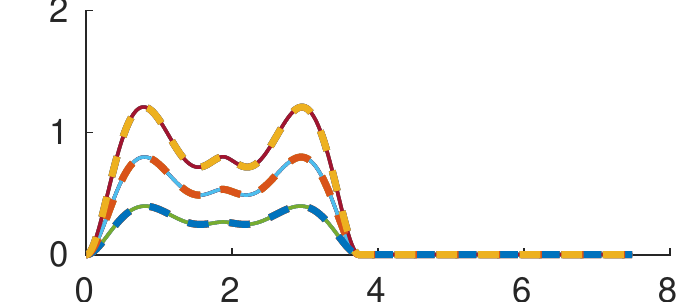}}
     ,(55,85)*{\text{Water Velocity}}
     ,(-5,50)*{\rotatebox{90}{\text{m/s}}}
     ,(80,25)*{\text{time (s)}}
   \end{xy}
   \end{minipage}
\hspace{.1cm}
   \begin{minipage}{.24\textwidth}
   \begin{xy}
    \xyimport(100,100){\includegraphics[width=3.8cm, trim=75 75 340 630, clip=true]{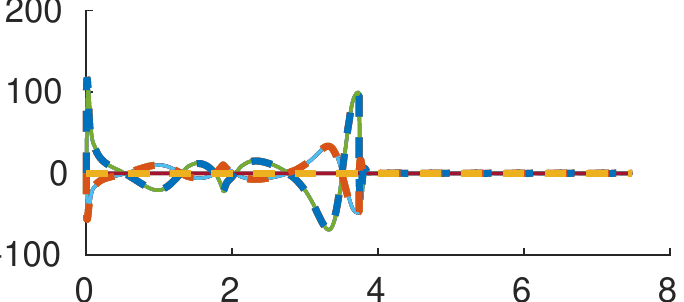}}
     ,(55,85)*{\text{Water Body Rates}}
     ,(-5,50)*{\rotatebox{90}{\text{deg/s}}}
     ,(80,25)*{\text{time (s)}}
   \end{xy}
   \end{minipage}\\

\vspace{.15cm}

   \begin{minipage}{.24\textwidth}
   \begin{xy}
    \xyimport(100,100){\includegraphics[width=3.8cm, trim=75 75 340 630, clip=true]{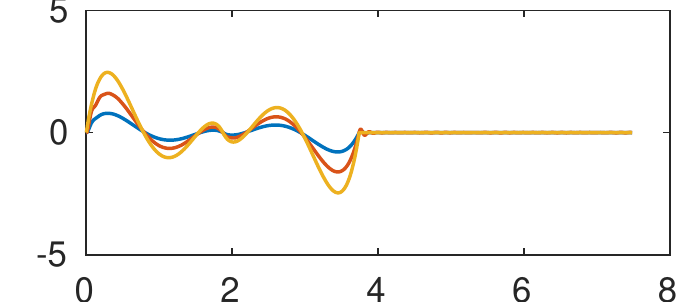}}
     ,(55,85)*{\text{Air Acceleration}}
     ,(-5,50)*{\rotatebox{90}{\text{m/s$^2$}}}
     ,(80,25)*{\text{time (s)}}
   \end{xy}
   \end{minipage}
\hspace{.1cm}
   \begin{minipage}{.24\textwidth}
   \begin{xy}
    \xyimport(100,100){\includegraphics[width=3.8cm, trim=70 75 340 630, clip=true]{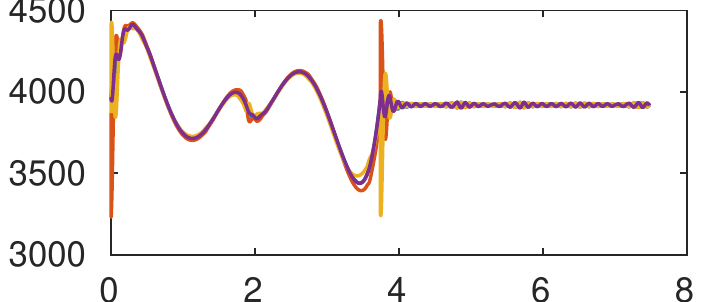}}
     ,(55,85)*{\text{Air Motor RMP}}
     ,(-5,50)*{\rotatebox{90}{\text{rpm}}}
     ,(80,25)*{\text{time (s)}}
   \end{xy}
   \end{minipage}
   \hspace{.15cm}
   \begin{minipage}{.24\textwidth}
   \begin{xy}
    \xyimport(100,100){\includegraphics[width=3.8cm, trim=75 75 340 630, clip=true]{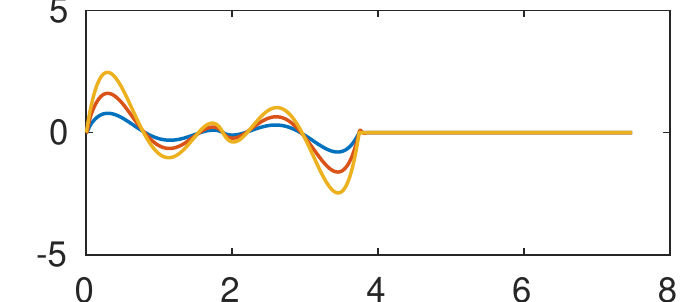}}
     ,(55,85)*{\text{Water Acceleration}}
     ,(-5,50)*{\rotatebox{90}{\text{m/s$^2$}}}
     ,(80,25)*{\text{time (s)}}
   \end{xy}
   \end{minipage}
\hspace{.1cm}
   \begin{minipage}{.24\textwidth}
   \begin{xy}
    \xyimport(100,100){\includegraphics[width=3.8cm, trim=75 75 340 630, clip=true]{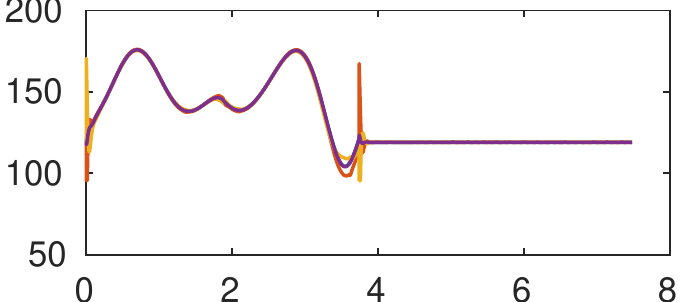}}
     ,(55,83)*{\text{Water Motor RPM}}
     ,(-5,50)*{\rotatebox{90}{\text{rpm}}}
     ,(80,25)*{\text{time (s)}}
   \end{xy}
   \end{minipage}\\

\vspace{.15cm}

   \begin{minipage}{.24\textwidth}
   \begin{xy}
    \xyimport(100,100){\includegraphics[width=4cm, trim=70 73 340 630, clip=true]{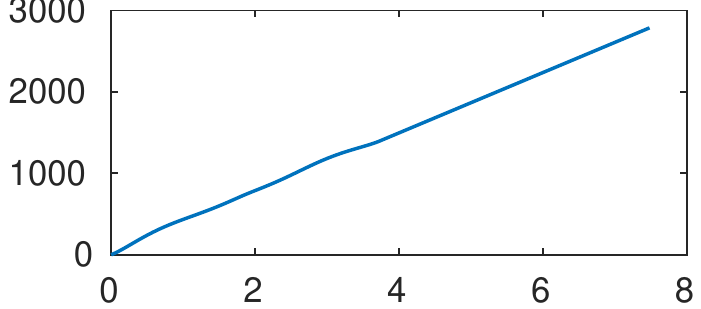}}
     ,(55,83)*{\text{Air Energy}}
     ,(-5,50)*{\rotatebox{90}{\text{Joules}}}
     ,(80,25)*{\text{time (s)}}
   \end{xy}
   \end{minipage}
   \hspace{.5cm}
   \begin{minipage}{.24\textwidth}
   \begin{xy}
    \xyimport(100,100){\includegraphics[width=7cm, trim=0 0 0 0, clip=true]{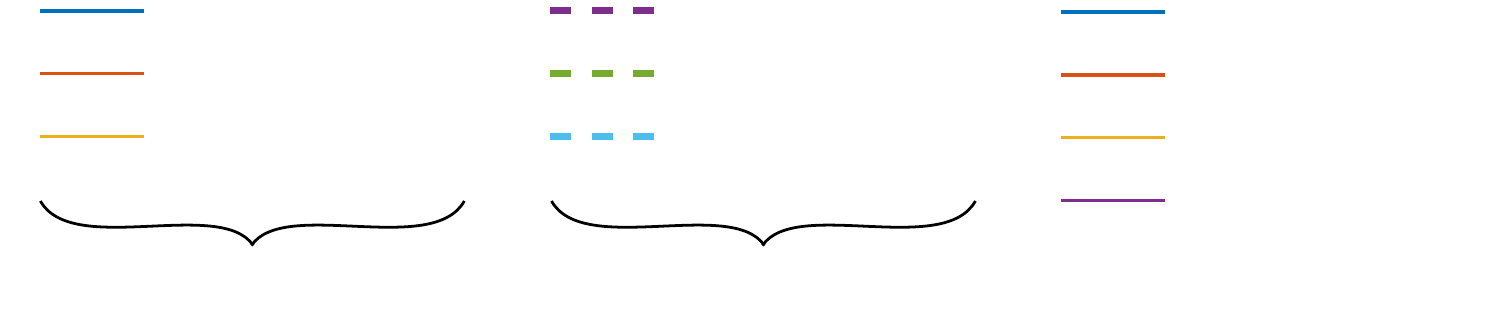}}
     ,(13,99)*[r]{\text{\footnotesize $x$, $\dot{x}$, $\ddot{x}$, $\psi$, $\dot{\psi}$}}
     ,(13,80)*[r]{\text{\footnotesize $y$, $\dot{y}$, $\ddot{y}$, $\theta$, $\dot{\theta}$}}
     ,(13,61)*[r]{\text{\footnotesize $z$, $\dot{z}$, $\ddot{z}$, $\phi$, $\dot{\phi}$}}
     ,(47,99)*[r]{\text{\footnotesize $x$, $\dot{x}$, $\ddot{x}$, $\psi$, $\dot{\psi}$}}
     ,(47,80)*[r]{\text{\footnotesize $y$, $\dot{y}$, $\ddot{y}$, $\theta$, $\dot{\theta}$}}
     ,(47,61)*[r]{\text{\footnotesize $z$, $\dot{z}$, $\ddot{z}$, $\phi$, $\dot{\phi}$}}
     ,(81,99)*[r]{\text{\footnotesize Motor $1$}}
     ,(81,80)*[r]{\text{\footnotesize Motor $2$}}
     ,(81,60)*[r]{\text{\footnotesize Motor $3$}}
     ,(81,41)*[r]{\text{\footnotesize Motor $4$}}
     ,(18,10)*{\text{\footnotesize Observed Values}}
     ,(52,10)*{\text{\footnotesize Target Values }}
   \end{xy}
   \end{minipage}
   \hspace{1.1cm}
   \begin{minipage}{.24\textwidth}
   \begin{xy}
    \xyimport(100,100){\includegraphics[width=4cm, trim=60 73 340 630, clip=true]{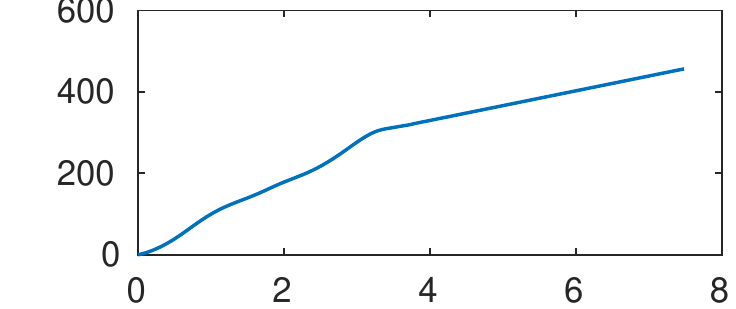}}
     ,(55,83)*{\text{Water Energy}}
     ,(0,50)*{\rotatebox{90}{\text{Joules}}}
     ,(80,25)*{\text{time (s)}}
   \end{xy}
   \end{minipage}

   \caption{\textit{ Left: Controller performance in air. Right: Controller performance underwater. Bottom-Center: legend.}}
   \label{fig:controller_kong}
\end{figure*}

\section{Appendix} \label{s:appendix}

\subsection{Procedurally Generated Environments}
Procedurally generated environments caves were made by first creating 2 sets of random 3D Perlin Noise \cite{Perlin} the size of the environment, which was chosen such that computation time was not excessive, yet environments are featured. Each set of noise encodes an angle $\theta$ or $\phi$ with a range of $\pi$ and $1.1\pi$ respectively. These ranges ensure that the caves do not fold back on themselves, and generally move across the entire environment. Parameters for number of bores $N_{bores}$, minimum and maximum length of bores $n_{min}$ and $n_{max}$, length of bore segment $l_{bore}$, and bore radius $R_{bore}$ are set beforehand, and chosen such that caves are likely to intersect and the environment remains concave. For each bore a random point is chosen at which the Perlin Noise creates a vector $(l_{bore},\theta,\phi)$ in spherical coordinates leading to the next point. This is repeated until either $n_{max}$ bore segments have been created or a vector leads to a point outside the bounds of the environment. The whole process is repeated until $N_{bores}$ bores with more than $n_{min}$ segments are found. Voxels within $R_{bore}$ of bore points are marked as free space (all else is obstacles). The water level is set to exactly half of the height of the environment. A random y-z and x-y plane are mapped as free space to make the environment dynamic.

\subsection{Illustration of Controller Performance}

Fig. \ref{fig:controller_kong} depicts an example of the position, velocity, acceleration, and motor RPM vs.\ time for both an air and underwater maneuver from (0,0,0) to (1,2,3).

\bibliographystyle{IEEEtran}
\bibliography{sample.bib}

\end{document}